\documentclass[sigconf,10pt,hyphens]{acmart}
\setcopyright{rightsretained} 

\hypersetup{draft}

\usepackage[ruled,vlined,linesnumbered,noend]{algorithm2e}
\usepackage{graphicx,wrapfig}
\usepackage{textcomp}
\usepackage{multirow}
\usepackage[utf8]{inputenc}
\usepackage{csquotes}
\usepackage{subcaption}
\usepackage{verbatim}
\usepackage{enumitem}
\usepackage{xcolor}

\usepackage[capitalize,noabbrev,nameinlink]{cleveref}

\usepackage{ulem}

\usepackage{tight}
\usepackage{tabularx}

\usepackage{tikz}

\usepackage{listings}
\usepackage{xcolor} 

\lstdefinestyle{mystyle}{
    commentstyle=\color{gray}\textit,
    keywordstyle=\color{blue}\bfseries,
    stringstyle=\color{purple},
    basicstyle=\tiny\ttfamily\linespread{.1}\selectfont,
    numbers=left,
    numberstyle=\tiny\color{gray},
    stepnumber=1,
    numbersep=1pt,
    backgroundcolor=\color{white},
    showspaces=false,
    showstringspaces=false,
    showtabs=false,
    frame=lines,            
    framesep=0mm,
    tabsize=1,
    captionpos=b,
    breaklines=true,
    breakatwhitespace=false,
    breakautoindent=true,
    postbreak=\mbox{\textcolor{red}{$\hookrightarrow$}\space},
    emphstyle=\color{red},
    morekeywords={True, False, None, while, if, else, world, sync_mode, client_port, fixed_delta_seconds, seed, edge_base, target_speed, num_lanes, edge_dt, search_dt, edge_sets_destination, spawn_position, destination, sensing, perception, localization, behavior, max_speed, overtake_allowed}, 
    literate={<<}{{\textless\textless}}{2}  
             {*}{{\char`\*}}{1}            
             {&}{{\char`\&}}{1}            
             {:}{{:\ }}{1}                
}

\crefformat{section}{#2§#1#3}
\crefformat{subsection}{#2\S#1#3}
\crefformat{subsubsection}{#2\S#1#3}

\newcommand{\TheName}{\textit{eCAV}\xspace}

\acmISBN{}
\acmConference[]{}{}{}

\acmDOI{}
\acmISBN{}
\acmPrice{}
\acmYear{}
\copyrightyear{}
\acmConference[]{}{}{}{} 
\acmBooktitle{}

 \def\@testdef #1#2#3{%
   \def\reserved@a{#3}\expandafter \ifx \csname #1@#2\endcsname
  \reserved@a  \else
 \typeout{^^Jlabel #2 changed:^^J%
 \meaning\reserved@a^^J%
 \expandafter\meaning\csname #1@#2\endcsname^^J}%
 \@tempswatrue \fi}

\begin{document}
\title{\TheName: An Edge-Assisted Evaluation Platform for Connected Autonomous Vehicles}

\baselineskip=11pt

\author{Tyler C. Landle}
\email{tlandle3@gatech.edu}
\affiliation{%
  \institution{Georgia Institute of Technology}
  \city{Atlanta}
  \state{Georgia}
  \country{USA}
}

\author{Jordan Rapp}
\email{jrapp7@gatech.edu}
\affiliation{%
  \institution{Georgia Institute of Technology}
  \city{Atlanta}
  \state{Georgia}
  \country{USA}
}

\author{Dean Blank}
\email{dean.blank@gatech.edu}
\affiliation{%
  \institution{Georgia Institute of Technology}
  \city{Atlanta}
  \state{Georgia}
  \country{USA}
}

\author{Chandramouli Amarnath}
\email{chandamarnath@gatech.edu}
\affiliation{%
  \institution{Georgia Institute of Technology}
  \city{Atlanta}
  \state{Georgia}
  \country{USA}
}

\author{Abhijit Chatterjee}
\email{abhijit.chatterjee@ece.gatech.edu}
\affiliation{%
  \institution{Georgia Institute of Technology}
  \city{Atlanta}
  \state{Georgia}
  \country{USA}
}

\author{Alexandros Daglis}
\email{alexandros.daglis@cc.gatech.edu}
\affiliation{%
  \institution{Georgia Institute of Technology}
  \city{Atlanta}
  \state{Georgia}
  \country{USA}
}

\author{Umakishore Ramachandran}
\email{rama@gatech.edu}
\affiliation{%
  \institution{Georgia Institute of Technology}
  \city{Atlanta}
  \state{Georgia}
  \country{USA}
}

\begin{abstract}
 As autonomous vehicles edge closer to widespread adoption, enhancing road safety through collision avoidance and minimization of collateral damage becomes imperative. Vehicle-to-everything (V2X) technologies, which include vehicle-to-vehicle (V2V), vehicle-to-infrastructure (V2I), and vehicle-to-cloud (V2C), are being proposed as mechanisms to achieve this saftey improvement.

Simulation-based testing is crucial for early-stage evaluation of Connected Autonomous Vehicle (CAV) control systems, offering a safer and more cost-effective alternative to real-world tests. However, simulating large 3D environments with many complex single- and multi-vehicle sensors and controllers is computationally intensive. There is currently no evaluation framework that can effectively evaluate realistic scenarios involving large numbers of autonomous vehicles.

We propose \TheName -- an efficient, modular, and scalable evaluation platform to facilitate both functional validation of algorithmic approaches to increasing road safety, as well as performance prediction of algorithms of various V2X technologies, including a futuristic Vehicle-to-Edge control plane and correspondingly designed control algorithms. \TheName can model up to 256 vehicles running individual control algorithms without perception enabled, which is $8\times$ more vehicles than what is possible with state-of-the-art alternatives.
With perception enabled, \TheName simulates up to 64 vehicles with a step time under 800ms, which is $4\times$ more and $1.5\times$ faster than the state-of-the-art OpenCDA framework.


 
\end{abstract}

\maketitle

\section{Introduction}

\par Autonomous Vehicles (AVs) require sensor information from a variety of sources to establish ground truth to properly navigate roads and infrastructure without human intervention. Recent research has focused on the area of cooperative perception and sensor data sharing as it relates to trade-offs between autonomous driving performance and local decision-making~\cite{tsukada_autoc2x_2020}. Decision-making of self-driving cars that solely rely on information harvested from local sensors is inherently limited, being unable to perceive objects occluded by other objects~\cite{emp}. In addition, local decision-making is suboptimal to account for traffic flow and safety~\cite{collab_manuevering_motivation}.


Prior art in evaluation frameworks for self-driving cars is at best good for \textit{single autonomous vehicle} simulations in the presence of V2X and collaborative perception~\cite{opencda,autocast,mosaic_cosim}. In addition, existing tools leave many unanswered questions regarding key pain points of today's AV evaluation needs, including support for evaluation of scenarios with large numbers of independent AVs~\cite{emp}. Existing tools provide incomplete solutions, such as removing key parts of the AV processing stack and replacing them with direct simulator information to support larger number of AVs, at 
the expense of simulation accuracy~\cite{CARLA_Traffic_Manager,autocast,opencda}. Evaluation frameworks with perception enabled in each vehicle client, wherein sensor data generated from a 
scenario generator (such as CARLA~\cite{CARLA}) is ingested and processed leading to control actions, cannot support simulations on the order of tens or hundreds of AVs~\cite{emp}.


Prior art is also limited in the type of interactions modeled between the vehicles.
For example, Schwarting, et al.~\cite{annurev-control-060117-105157} assume some level of control of the external locale over the decision-making of the ego vehicle via V2V communication. In the real world, there will be limited control over other vehicles, due to the number of human-controlled vehicles and heterogeneity of control systems on-board. Real-world systems must account for both trajectories produced by other self-driving vehicles, and predict the trajectories of human-controlled vehicles. Computational costs for tasks such as short and medium term motion planning are also high. 

In the most challenging traffic scenarios, the computational resources and data available locally at each individual autonomous vehicle are insufficient for optimized medium length motion planning augmented with \textit{beyond line of sight (BLOS)} information. While ample computational resources are within reach in the cloud, network latency limits preclude offloading such functionality to the cloud~\cite{edge_cloud_collab}. 
Given this challenges, prior research suggests \textit{edge} computing as a promising building block complementary to local (on-vehicle) or cloud-based processing, minimizing network latency while providing sufficient computational power to offload some of these tasks \cite{ecloud}. However, to evaluate such futuristic systems, new evaluation frameworks that support the modeling of such edge-based support are needed.

To address these limitations, we propose the \TheName platform, which makes the following contributions:
\begin{itemize}
    \item \textbf{A scalable evaluation platform that can be deployed in a distributed fashion to enable large-scale self-driving vehicle simulations}: \TheName provides a centralized programming interface to support performant large-scale, user-defined, multi-AV scenarios that can be deployed across multiple physical or virtual machines. \TheName facilitates the deployment, monitoring, and post-experiment measurement and analysis capabilities for a developer of a CAV system to answer many ``what if'' questions. 
    \item \textbf{A pluggable, generalizable edge interface where the 
    CAV system developer can define, execute, and evaluate scenarios that utilize the edge as a component in the CAV system}: 
    To facilitate future research on edge-assisted vehicle control at scale, \TheName makes it possible to simulate the execution of algorithms used in decision-making locally in the vehicle and in the edge. \TheName supports large-scale multi-vehicle simulations that can be used to evaluate the benefits and drawbacks of such algorithms in varied latency settings among the entities.
\end{itemize}


In Section \ref{sec:context}, we first explore the need for simulation frameworks within the AV space. We then review the requirements and limitations of the current-state-of-the-art in Section \ref{sec:requirements}.

\TheName's architecture, covered in Section \ref{simarch} along with specific implementation details in Section \ref{sec:impl}, was designed from the outset with an emphasis on performance. In addition to performance, we also emphasized usability, covered in Section \ref{sec:usability}. In Section \ref{sec:eval}, we demonstrate \TheName's scalability with microbenchmarks and end-to-end evaluation of up to 256 simulated vehicles running unique control stacks, deployed across 4 cloud-based virtual machines. With the same computational resources, \TheName supports up to 64 simulated vehicles with perception enabled in each vehicle client, wherein sensor data generated from the CARLA simulator is ingested and processed leading to control actions. We conduct a case study exemplar application to demonstrate how \TheName facilitates enables research for CAVs utilizing the edge.

\section{Context}
\label{sec:context}

\par For reasons of both cost and safety, simulated testing is an essential tool for AV and CAV development. Advanced simulators critically allow continuous development without wear and tear on a physical vehicle and constant human supervision, and freedom to test novel control algorithms without fear of putting other vehicles or pedestrians at risk. Simulated vehicles can also cover orders of magnitude more miles than physical AVs, and be tested under virtually infinite traffic scenarios.
Waymo's vehicles have driven 20+ million physical miles and 15+ billion simulated miles as of 2021 \cite{waymo2020}.
Simulation-based approaches to testing are typical based on a larger analysis of testing within the AV space~\cite{testing_where_are_we}.


Due to CAV simulation necessity, significant effort is being put into the development of advanced simulators that can faithfully render immersive and photo-realistic 3D environments, leveraging state-of-the-art game engines such as UnrealEngine4 (CARLA)~\cite{CARLA} and Unity (LGSVL)~\cite{lgsvl} that offer powerful rendering pipelines and a rich set of standard APIs for environment modeling. 
In a survey of state-of-the-art simulators, Zhou et al.~\cite{surveyofavsims} discuss CarSim, CarMaker, PreScan, Air-Sim, CARLA, LGSVL, and Matlab/Simulink as examples of the most commonly used simulators. CarSim and CarMaker focus on vehicle dynamics, while newer simulators put a greater emphasis on testing advanced perception, localization (i.e., determining position and orientation on the map), and planning algorithms---i.e.,  features that represent the critical components of an AV control system. 





\section{Requirements}
\label{sec:requirements}



In \cref{table:state_of_the_art}, we have captured the essential attributes for a large-scale evaluation framework for CAVs. We have arrived at these attributes based on the expected evolution of CAVs as reported by the National Highway Safety Administration~\cite{car-timeline}, and on relevant literature that have tried to identify the functional requirements for such frameworks~\cite{surveyofavsims, surveyofselfdriving}. \TheName, described in the subsequent sections, is designed to meet all these attributes.

\subsection{Comparing Evaluation Frameworks} \label{stateoftheart}

\cref{table:state_of_the_art} summarizes the capabilities of the state-of-the-art evaluation frameworks (namely, AutocastSim \cite{autocast} and Open\-CDA \cite{xu2021opencda}) with respect to the attributes identified.
We note that none of the state-the-art evaluation frameworks offer support for evaluating edge-assistance in CAVs.  We elaborate on the specific needs for edge-assistance in \cref{controlplanerequirements}.

We use OpenCDA as our main comparison point in the rest of the paper. OpenCDA is an open-source, Python-based evaluation framework \cite{xu2021opencda}. It offers cooperative-driving automation (CDA) in the form of platooning as well as a suite of standard modules for regular automated driving including perception, planning, and control. OpenCDA meets many of the core requirements for an evaluation framework, but it is single-threaded and therefore sequential execution  offers limited scalability, especially for perception-enabled scenarios where GPU compute resource requirements are high.

\preto\tabular{\setcounter{magicrownumbers}{0}}
\newcounter{magicrownumbers}
\newcommand\rownumber{\stepcounter{magicrownumbers}\arabic{magicrownumbers}}
\begin{table}[ht]
\footnotesize
\begin{tabular}{ |p{3.3cm}| >{\centering\arraybackslash}p{1.4cm}|>{\centering\arraybackslash}p{1.2cm}|>{\centering\arraybackslash}p{1cm}|  }
 \hline
 \multicolumn{4}{|c|}{State-of-the-Art Evaluation Frameworks} \\
    \hline
 \textbf{Attribute}     & \multicolumn{1}{|c|}{AutoCastSim} & \multicolumn{1}{|c|}{OpenCDA} & \multicolumn{1}{|c|}{\TheName} \\
    \hline
  {\textit{R1}}: Stable and offering reliable, repeatable results & yes & yes & yes\\
 \hline
  {\textit{R2}}: Centralized programming model to clearly specify scenarios & yes & yes & yes\\
  \hline
  {\textit{R3}}: Supports Independent Algorithm Input & yes & yes & yes\\
  \hline
  {\textit{R4}}: Visualization of a Photo-Realistic Environment & yes & yes & yes\\
  \hline
  {\textit{R5}}: Precise, Accurate Vehicle Dynamics and Control & yes & yes & yes\\
  \hline
  {\textit{R6}}: Linear Scaling of Scenario Performance with number of physical machines & \bf{no} & \bf{no} & \bf{yes}\\
  \hline
  {\textit{R7}}: Portability to support heterogeneous computing resources & \bf{no} & \bf{no} & \bf{yes}\\
  \hline
  {\textit{R8}}: Supports Large Number of Independent Vehicles & \bf{no}  & \bf{no} & \bf{yes}\\
  \hline
  {\textit{R9}}: Accurate modeling of network characteristics & \bf{no} & \bf{no}  & \bf{yes}\\
  \hline
  {\textit{R10}}: \textbf{Edge Support} & \bf{no} & \bf{no}  & \bf{yes}\\
 \hline
\end{tabular}
\caption{State-of-the-Art CAV Evaluation Frameworks Comparison}
\label{table:state_of_the_art}
\vspace{-4mm}
\end{table}

\subsection{Edge-Based Control Plane Requirements} \label{controlplanerequirements}

Edge-based control planes represent an ongoing area of research for next-generation CAVs \cite{ecloud}. Use of the edge can help meet the tight latency requirements of CAVs, and simplify many existing control approaches when dealing with CAVs. The edge control plane must be able to handle: 
\begin{enumerate}
    \item Variable latency in vehicle response times due to last mile connectivity
    \item Missed responses from vehicles due to loss of connectivity
    \item Missed acknowledgements/receipt-of-data due to vagaries of the network 
    \item Vehicles entering and exiting the edge node locale
\end{enumerate}

Latency and networking are critical, as the edge control plane must process inputs and act on them quickly enough to provide useful information back to vehicles. 
The edge is particularly well-suited for gathering BLOS data and data about human-operated vehicles and disseminating them to the affected AVs. To properly evaluate the performance of edge-based control planes, an evaluation framework must:
\begin{enumerate}
    \item Have a plug-and-play interface for rapid deployment and testing of new control algorithms
    \item Offer modular support for modeling networking characteristics 
    \item Account for heterogeneity (in terms of types of data, quality, and timeliness) of inputs and responses
\end{enumerate}

In the next section, we will explore how the architecture of \TheName was designed to meet these requirements.

\section{Architecture of \TheName} \label{simarch}

\begin{figure*}[htbp]
  \centering
  \includegraphics[width=\textwidth]{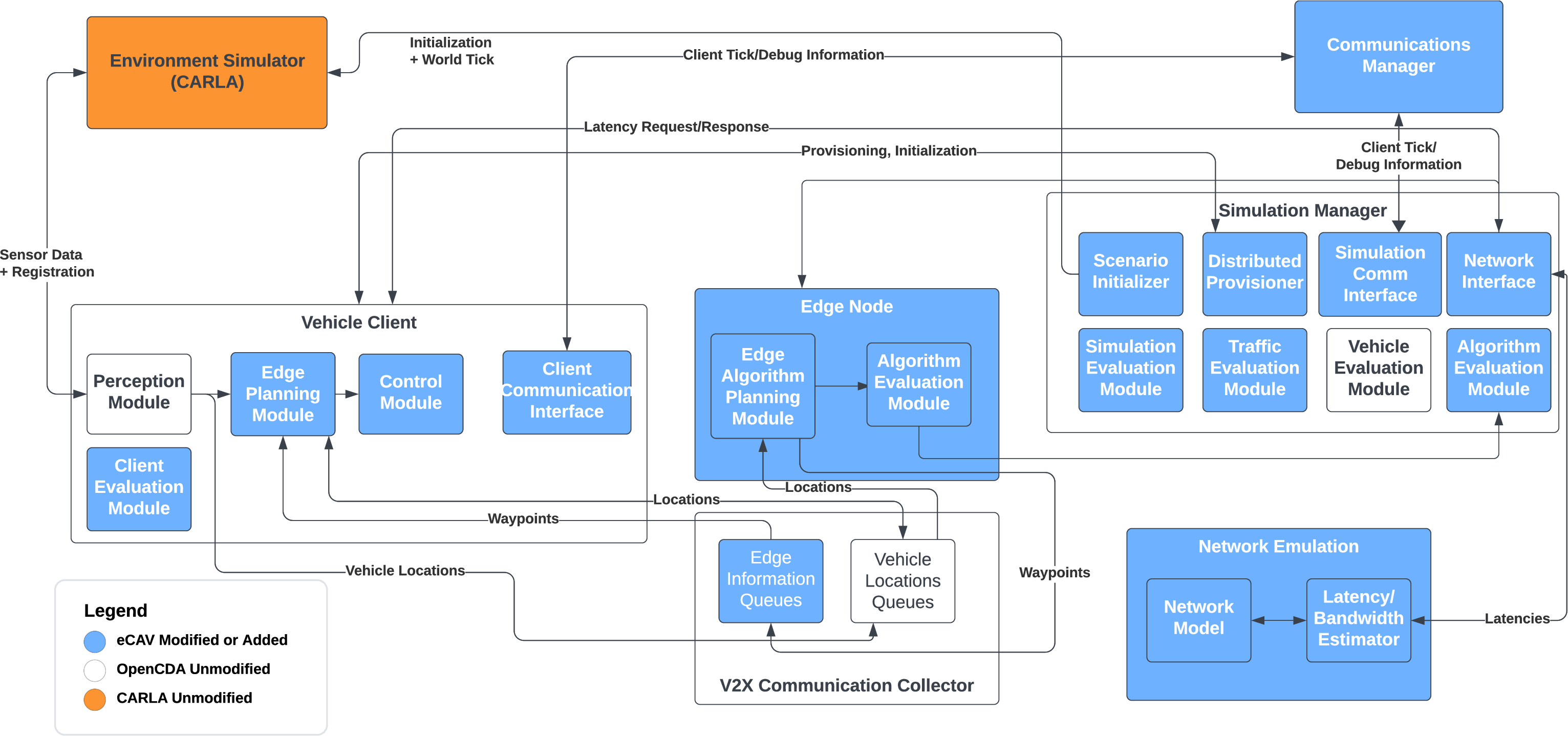}  
  \vspace{-8mm}
  \caption{\TheName Architecture.}
    \vspace{-2mm}
    \Description[]{}
    \label{fig:sim_arch}
\end{figure*}


Given the main goal of providing scalable evaluation of large-scale CAV scenarios, the architecture of \TheName was designed explicitly to meet the core requirements of performance-focused asynchronous communication, and distributed and parallel execution of all the actors (vehicles, edge, and the interactions among them). 

Much of the novelty in \TheName is embodied in the system techniques that help us to achieve the scalable performance goal for evaluating CAV systems while preserving the accuracy of the results. These include: (a) containerization and parallel execution of the control loops of the vehicle clients; (b) minimizing the communication overhead for all the inter-actor interactions; and (c) efficient resource sharing of GPUs for perception processing of the vehicle clients co-located on the same physical node. 


The individual constituent components were designed with modularity---and through that, future extensibility---in mind, to accommodate various areas of CAV research. 
\TheName's architecture, shown in \cref{fig:sim_arch}, consists of seven main entities: 
\begin{enumerate}
     \item \textbf{Simulation Manager:} Coordinates the interactions among all entities within a scenario.
     
      \item \textbf{Vehicle Client:} Models the control action within each independent modeled AV. 

     \item \textbf{Communication Manager:} Coordinates synchronized communication between the individual vehicle clients and the simulation manager.

    \item \textbf{V2X Communication Collector:} Acts as a central repository for collecting time-indexed information from the edge and individual vehicle clients. We rely on Open\-CDA \cite{xu2021opencda} to implement the functionality of collecting this information from the vehicle clients.

     \item \textbf{Environment Simulator:} Simulates client physics, streets, layout, generates waypoints, and sensor data. We use CARLA~\cite{CARLA} to serve this functionality.
     
     \item \textbf{Edge Support:} Provides a pluggable interface that allows incorporation of new edge-deployed control algorithms by domain experts. 

     \item \textbf{Network Emulation:} Provides a pluggable interface that  models the characteristics of connectivity between the edge node and the vehicle clients.
     
\end{enumerate}


The separation of responsibilities from a monolith to distinct components in the simulation manager, parallelization of individual vehicle clients, and the structure of the communication manager are the primary ways in which \TheName meets both the principal requirement of scalability---\textit{R6}---as well as supporting heterogeneous computing resources---\textit{R7} in \cref{table:state_of_the_art}. The dedicated edge node and network emulation component modules provide support for edge control plane development, i.e., \textit{R10}.


\subsection{Simulation Manager}

\par The role of the simulation manager is to set up, initialize, and manage the simulation environment. By moving from a monolithic approach where the simulation executed as a single process to a distributed framework where the simulation manager serves as a central orchestrator was a key step in achieving scalability and fulfilling \textit{R2} and \textit{R7} in \cref{table:state_of_the_art}.

\subsubsection{Scheduling}

In \TheName, the algorithms that run on the edge node are executed on some available host hardware (e.g., a server-grade CPU) outside of the rest of the entities shown in Figure \ref{fig:sim_arch}. As a result, synchronizing the simulated time with the edge algorithm's execution time requires time reconciliation (i.e., milliseconds of real time to simulation time steps).

Imagine a simulation where a car needs to make a right turn at a busy intersection. During the time step, the edge node may send a command message to tell the car when to turn. To properly simulate the real world, we may want the message to get there after 10ms rather than the current step.

The simulation manager serves as the intermediary between the clients and the environment simulator (CARLA) to ensure that the simulation is timing accurate. It ensures that messages from the vehicles are delivered to the edge node in the correct simulation time step.

If edge-assisted CAV is modeled, once the inputs from the vehicle clients and the environment simulator are received, the edge node performs the user-defined algorithm (described in \cref{sec:algo}) that fuses the inputs and generates the outputs for each of the vehicle clients and the environment simulator. 
The 
outputs are sent to the V2X Communications Collector for delivery in the correct simulation time step to the environment simulator and the affected vehicle clients.

\subsubsection{Collecting and Aggregating Evaluation Metrics}
As can be seen in \cref{fig:sim_arch}, each of the actors (such as vehicle clients and the edge node) gather metrics of interest during the course of the simulation.  To ensure scalability (\textit{R6}) it is important that the communication from the actors to the simulation manager to collect and aggregate such metrics is not in the critical path. A key optimization to this end is distributing the metric collection in each actor, and batching the communication to the simulation manager only at the end of the simulation run.  This optimization ensures \textit{R6} and \textit{R8}, while not compromising on \textit{R1} and \textit{R2}.  At the same time, the collection and aggregation of all the evaluation metrics (client metrics, edge metrics, algorithmic metrics, network metrics, and scenario metrics) in one place, namely, the simulation manager provides researchers with a direct means of generating charts and graphs that measure scenario and simulation deployment performance.

\subsection{Vehicle Clients}

\textit{R8} requires individual vehicle clients to exist as standalone processes that asynchronously communicate with the simulation manager to achieve high-performance and scalability. 
We opt to implement that capability using container technology, thus jointly achieving \textit{R7} and \textit{R8}. 

\par All the actors in \TheName---vehicles, pedestrians, and sensors including the edge (detailed in Section \ref{sec:edge})---can send commands to CARLA (the environment simulator) and receive relevant requested information. Calls to CARLA's APIs are blocking and also can result in substantial data transfer; for example, the sensory information---RGB camera feeds, LIDAR point clouds, GPS data, and acceleration data from the Inertial Measurement Unit (IMU)---that helps control actuation of the vehicle can be quite large, especially in a perception-enabled scenario. 
Parallelization of the vehicle client is essential for meeting \textit{R6} and \textit{R8}, allowing the clients to make parallel calls to CARLA as well as to process the received information for their respective control algorithms.

\subsubsection{Integrating Edge-Node Support in Individual AV Control Decisions}
The pluggable algorithms that are housed in the edge node for fusing information from multiple vehicles and disseminated to the vehicle clients provide visibility to an individual vehicle beyond its line of sight.  Such information can be considered \textit{super-sensor} input to the vehicle control algorithm.  However, there are latency considerations for receiving such information from the edge node to the vehicle client and making timely decisions. The vehicle clients entity in \TheName include support for over-riding the edge node super-sensor input and using local sensor data in their control algorithms based on the staleness of the data they receive.


\subsubsection{Vehicle Planning And Control}

We make use of OpenCDA's ``ego''-vehicle planning and control algorithms, which have been shown to meet \textit{R1}, \textit{R3}, and \textit{R5} \cite{xu2021opencda}. The algorithms are themselves intended to interact with CARLA's API and are designed in a modular way to be easily modified or replaced. 

\subsection{Communication Manager}


\par The communication manager is key to \TheName's ability to meet \textit{R6}, \textit{R7}, and \textit{R8}. As the communication manager is entirely responsible for handling all network-based intra-simulation communication, it is able to manage heterogeneous computing resources \textit{R7} through the use of a hardware and programming-language agnostic communication paradigm. 

\begin{figure}[htbp]
  \centering
  \includegraphics[width=\columnwidth]{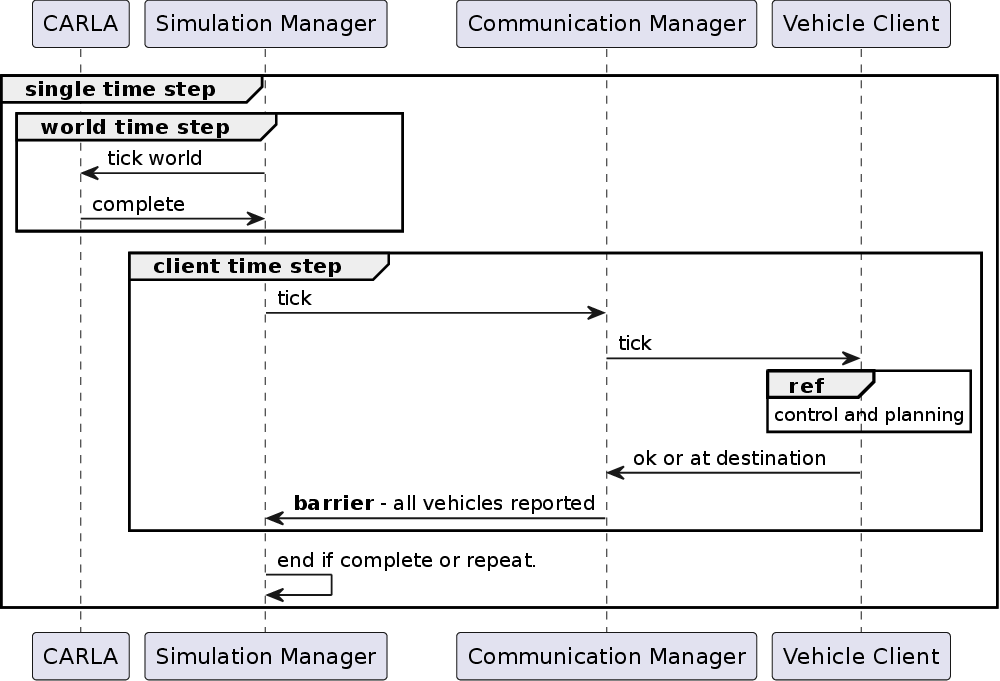}  
   \caption{Intra-Simulation Communications}
    \Description[]{}
   \label{fig:comms_arch}
\end{figure}

\subsubsection{Intra-Simulation Communication Protocol}

\par To facilitate communication between the simulation manager and individual vehicle clients, we used an event-based TCP communication paradigm. We utilize a parallel, multi-threaded, and asynchronous push-pull communication architecture to realize \textit{R6} and \textit{R8} in \cref{table:state_of_the_art}. A polling approach does not scale well to hundreds of clients with bursty traffic due to idle time, so a push-pull architecture was selected to handle intra-simulation communications. Generic events---such as initiating or completing a client tick---are pushed to lightweight servers running in the simulation manager and individual vehicle clients. More substantial data payloads or data that are only applicable to a given process, such as the specific waypoints determined by the edge for an individual vehicle, are instead selectively pulled from the communication server by the requesting process. 

\par Figure \ref{fig:comms_arch} depicts the network communications during a single time step in \TheName{}. While vehicle clients remain idle awaiting an event push from the simulation manager, the simulation manager processes data and instructs CARLA to advance the simulation, updating physics and world-state models based on the previous step's inputs. The simulation manager then directs the communication manager to initiate the next vehicle client time step. The simulation manager then enters an idle state, awaiting an event push indicating the completion of the current step by all vehicles, while they execute planning and control algorithms. 








\subsection{V2X Communication Collector}

The V2X Communications Collector collects time-indexed information from the edge and individual clients. This information includes sensor data, control signals, and other data relevant to the simulation.
The V2X Communications Collector acts as a central repository for all time-sensitive data in the simulation. It allows vehicles and the edge to access the most up-to-date information, even when they have different network latencies, as the host threads executing their functionality can be located on different machines of a cluster during simulation. This feature is essential for ensuring that the simulation is realistic and accurate.

Data from the clients to the edge also passes through the V2X Communications Collector, simplifying the programming model for programming the clients. The network model (described in Section \ref{subsec:network}) within \TheName allows clients to account for the network latency among the interacting entities commensurate with the physical characteristics of the network a user wishes to model for the simulated environment.

We use a lockless architecture to handle the V2X Communication Collector queues to realize scalability when many different clients need to access waypoints with applied network latency so that we can realize \textit{R6}, \textit{R8} and \textit{R9}, i.e., network modeling while still supporting scale with low overhead.

\subsection{Environment Simulator - CARLA}

\par CARLA~\cite{CARLA}, an open-source autonomous driving simulator, provides a photo-realistic environment with accurate physics modeling for AVs to operate in---meeting \textit{R4} and \textit{R5}. CARLA works in a client-server model, with the server controlling the simulation and providing sensor data to the clients. The environment itself is composed of 3D models of buildings, vegetation, and traffic signs. CARLA is built on top of Unreal Engine 4 (UE4), which provides physics, rendering, and basic NPC logic.

\par To ensure consistent simulation time, we run CARLA in synchronous fixed time-step mode, where the CARLA simulator waits until a designated actor ticks the simulation before processing the next frame. The simulation manager determines the simulation time tick, ensuring that every simulation time step is a constant simulated time interval.

\subsection{Edge Node Support for Coordinated Control}
\label{sec:edge}

\begin{figure}[htbp]
  \centering
  \includegraphics[width=\columnwidth]{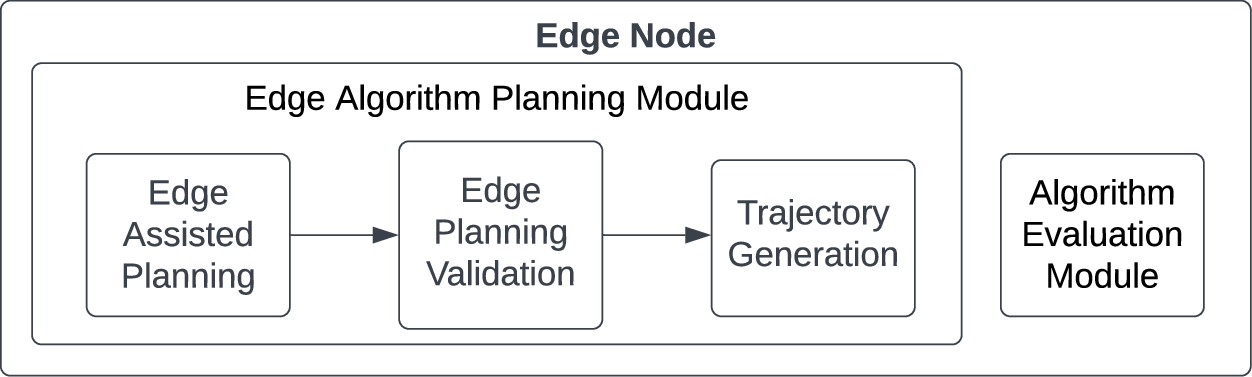}  
   \caption{Edge Node Zoomed In}
   \Description[]{}
   \label{fig:node_arch}
\end{figure}

To integrate the edge node support into \TheName, the edge node functions as an actor, similar to any other vehicle client. Its inputs are sensor signals from  the vehicle clients and its outputs are the proposed control actions that are sent back to clients, with the aim of improving traffic flow. The edge node, shown in \cref{fig:node_arch}, consists of three main components: edge assisted planning, the edge planning validation, and the algorithm evaluation module.

\subsubsection{Edge Assisted Planning}

 The edge node planning module runs actual algorithms intended to enhance autonomous traffic flow while maintaining safety of the autonomous vehicles. These algorithms can range from co-perception systems to enable sharing of information beyond an individual vehicle's line of sight to traffic management and lane selection systems that suggest optimal lane assignments or traffic maneuvers. The algorithms' outputs are collected by the edge node and used to send commands and/or provide \textit{super sensor} input to vehicle clients that are in the geographical vicinity of one another for use alongside local sensor data for vehicle control.

 A typical workflow of an algorithm that does route planning for the vehicles in its purview is shown in \cref{fig:node_arch}. It proceeds in three steps: edge assisted planning, edge planning validation, and trajectory generation. In the edge assisted planning step, the edge algorithm calculates waypoints for each vehicle client under its purview to follow. Following this step, the edge algorithm validates each waypoint against the road topology and the waypoints of other vehicles on the road to avoid collisions and faulty waypoints. 
 %
 The final step is to generate the suggested trajectories as waypoints for the vehicles that are in a given locale, which are then communicated to the vehicle clients. We demonstrate an example in Section \ref{sec:eval:usability}.

 We compute only the waypoints to be sent to the vehicle clients on the edge node so that we can reduce packet size and parallelize the planning process without reducing the accuracy of the simulation, tying back to \textit {R6}, \textit{R8} and \textit{R10}.

 \subsubsection{Edge Planning Validation}

In the edge planning validation module, we validate the positional waypoints generated by the edge planning module to ensure that they are valid waypoints for the vehicle clients to use. We also ensure no collisions or violations are computed by the edge assisted planning before generating trajectories.

 \subsubsection{Algorithm Evaluation Module}

 %
 The developer of co-perception algorithms running on the edge would want to measure metrics of interest such as running time of the algorithm as a function of number of vehicle clients. \TheName provides an extensible framework for collection of such metrics of interest. Additionally, it provides an interface to plot the collected metrics against each other at the end of a simulation run. In order to realize \textit{R10}, the edge must be able to collect useful metrics. Similar to the rest of our evaluation module, metrics are collected in bulk at the end of the simulation to improve runtime performance, satisfying \textit{R6}, \textit{R8}, and \textit{R10}.



\subsection{Network Emulator}
\label{subsec:network}

A \textit{pluggable} network model takes into consideration the location of the clients and edge nodes and the characteristics of the physical layer for communication among these entities. The existing model estimates the latency for each communication event, which the vehicle clients use to request the necessary information from the V2X Communications Collector to satisfy \textit{R9}.

\section{Implementation}
\label{sec:impl}

\par We implemented \TheName by integrating the architectural components presented in Section \ref{simarch} into OpenCDA version 0.1.1. We leverage the underlying individual vehicle management tools, such as behavior control, perception, and localization provided by OpenCDA, and preserve all existing functionality of its existing sequential-execution framework with minimal code changes. We extend it to support large-scale, distributed, and asynchronous simulations with an additional edge control plane and a simulated network model.

\subsection{Containerization}


\par Each vehicle process runs as an individual container, which provides a dedicated virtual environment with its own GIL. This allows Python processes to be parallelized even on a single compute node, without the limitation of a single GIL across all containers \cite{python_gil}.

\subsubsection{Perception}

To access GPU computing resources from the perception container, we used the nVidia Docker runtime~\cite{nvidia_docker2}. The nVidia runtime uses nVidia's CUDA framework for GPU computing. 
Because individual vehicles need not share approaches to control, not all vehicles need access to GPU resources. 



\subsection{Distribution}


\par CARLA's distributed communication framework uses HTTP traffic over UDP and supports \TheName-enabled distributed simulation without any modification. Individual vehicle clients and the simulation manager communicate directly with the CARLA server.

\par We used gRPC and protobuf  to communicate between the simulation manager and individual vehicle clients. The gRPC server is written in C++ to allow parallel communication, which is not possible with a Python-based server. We used gRPC's asynchronous callback architecture with the Python \texttt{asyncio} library.


\par Protobuf offers convenient multi-language communication between the individual Python processes (Simulation Manager, Vehicle Clients, and Edge) and the C++ communication manager. It also provides a framework for serializing various data structures allowing for centralized evaluation of distributed vehicle client data.

\section{Using \TheName}
\label{sec:usability}

Individual scenarios are defined in {YAML} files and a dedicated Python script. The \textit{scenario configuration YAML} file contains specifications for the simulated world, vehicle clients, and various control modules. A \textit{scenario definition} Python script contains the specific implementation details, in particular managing the syncing of various time-steps. A sample scenario definition script is shown below:

\begin{lstlisting}[language=Python, caption=Sample scenario definition script., label=lst:python_sample]
flag = True
waypoint_buf = []
world_time = 0
while flag:
    scenario_manager.tick_world()
    world_time += world_dt
    if world_time % edge_dt == 0:
        world_time = 0
        edge.update_information()
        waypoint_buf = edge.run_step()
        scenario_manager.push_waypoint_buffer(waypoint_buf)
        flag = scenario_manager.broadcast_message(ecloud.Command.PULL_WAYPOINTS_AND_TICK)
    else:
        flag = scenario_manager.broadcast_tick()
\end{lstlisting}

\TheName allows for a single locus of control over vehicle clients using the edge, which is sending the vehicle clients waypoints. Waypoints are defined as a tuple {{Location, Orientation}} which defines the 6 degrees of freedom location of which a vehicle client should set their next road destination. The vehicle clients each currently use a locally generated buffer of waypoints to direct the movement of the vehicle in the simulated world. \TheName provides an interface into the vehicles to (1) accept waypoints from the cooperative control plane and (2) validate and discard if they are not valid.

\subsection{World Configuration YAML Parameters}

The \textit{world configuration YAML} sets specifications for the simulated world, such as the fixed time step, whether we are running in synchronous mode, the CLARA client port, and other CARLA specific configuration parameters. An example configuration file snippet is shown below:

\label{world_config_yaml}
\begin{lstlisting}[caption=World configuration snippet., label=lst:world_config_yaml]
world:
    sync_mode: true
    client_port: 2000
    fixed_delta_seconds: &delta 0.03
    seed: 10
\end{lstlisting}

\subsection{Edge Specific YAML Configuration}

The \textit{edge configuration YAML} file specifies configuration parameters for the edge node, such as target speed, maximum capacity of cars the edge can support, and edge control loop time. Algorithm-specific configuration parameters can be added to this file to support other necessary parameters:

\label{edge_config_yaml}
\begin{lstlisting}[caption=Edge Configuration Snippet., label=lst:edge_config_yaml]
 edge_base: &edge_base
        target_speed: 55 
        num_lanes: 4
        edge_dt: 0.200
        search_dt: 2.00
        edge_sets_destination: true 
\end{lstlisting}

\subsection{Vehicle Specific YAML Configuration}

The \textit{Vehicle YAML configuration file} allows the user to configure specific vehicle parameters: 

\label{vehicle_config_yaml}
\begin{lstlisting}[caption=Vehicle Config Snippet., label=lst:vehicle_config_yamll]
    <<: *vehicle_base
        spawn_position: [-19.52, -19.29, 0.3, 0, 180, 0]
        destination: [-147.79, -12.29, 0]
        sensing:
          <<: *base_sensing
          perception:
            <<: *base_perception
            localization:
              <<: *base_localize
          behavior:
            <<: *base_behavior
            max_speed: 100
            overtake_allowed: true
\end{lstlisting}
\section{Evaluation}
\label{sec:eval}


We now evaluate \TheName{} to validate the following hypotheses:

\begin{enumerate}
\item \TheName{} improves scalability and performance compared to existing state-of-the-art CAV simulator frameworks.
\item Our extensions required to make simulation scalable and modular do not impact the accuracy of the simulation and have minimal effects on the performance of small-scale simulations.
\item \TheName{} is a useful and versatile tool for conducting evaluations of edge-assisted CAV algorithms.
\end{enumerate}

We evaluate \TheName along the following axes:

\begin{itemize}
\item Scalability: the time it takes to run a simulation with different numbers of AVs.
\item Performance: the step time that can be achieved with different numbers of AVs.
\item Accuracy: the proximity of \TheName simulation results to OpenCDA simulation results. 
\item Usability: as this metric is hard to quantify, we demonstrate \TheName's usability with a case study of an exemplar algorithm for assisting AVs using the edge. 
\end{itemize}

\subsection{Scalability Metrics}

In these experiments, step time is the time it takes the simulator to run one synchronized 50ms timestep (i.e., 50ms of the simulated environment's time), which is openCDA's default value to ensure stability of the simulated AVs' control loops and CARLA's physics engine. In each simulation time step, there are two main phases that make up a single time tick. Each phase must complete before the next phase starts. The two phases are:

\begin{enumerate}
    \item \textbf{Client Step Time} - time for each client to run its control processing loop and return. 
    \item \textbf{World Step Time} - time for the CARLA simulator to "tick" the environment simulator based on client vehicle inputs, and send relevant information back to all clients. 
\end{enumerate}

 All clients must complete their client step before the environmental simulator can update its state with the relevant inputs from the clients in the following world step phase. 
 \TheName parallelizes the client step time to enable simulation scalability with the number of simulated AVs. 
 Therefore, while we show how world step time scales with the number of clients, most of our evaluation focuses on client step time.
 Total simulation time is defined by the number of simulation steps executed; in turn, each simulation step's duration is dictated by the sum of the two aforementioned phases.
 
\subsection{Experimental Setup}

All experiments experiments are run within the environment of CARLA, deployed on Microsoft Azure Public Cloud Virtual Machines (VM) \cite{Azure}. 
To evaluate CARLA and state-of-the-art framework scalability, we simulate traffic scenarios on the CARLA Town06 map and a custom highway map generated by the developers of OpenCDA \cite{xu2021opencda}.
Our evaluation covers four different scenarios: simulation deployment on a single machine or multiple machines, with and without perception enabled.

\subsubsection{Single-Node without Perception}

We utilize the Standard NV12s v3 \cite{NVS_V3} series GPU-supported VMs to run the CARLA simulator, whereby each VM contains 12 cpu cores, 112 GiB of RAM, and 320 GB of premium SSD storage. Each VM also has access to one core of a Tesla M60 \cite{M60} GPU and 8 GiB of GPU memory. One Standard D64s v5 \cite{Dsv5} is used to run the Vehicle Clients and \TheName's Simulation Manager. This VM contains 64 vCPUs with hyperthreading support and 256 GiB of RAM, with no GPU support.

\subsubsection{Single-Node with Perception Enabled}

We utilize the Standard NV12s v3 \cite{NVS_V3} series GPU-supported VMs to run the CARLA simulator. We use the Standard ND40rs v2~\cite{NdM} to run the Vehicle Clients and \TheName's Simulation Manager. Each VM contains 40 vCPUs with hyperthreading support and 672 GiB of RAM, as well as eight Tesla V100 GPUs. 

\subsubsection{Multi-Node without Perception}

We utilize the Standard NV12s v3 \cite{NVS_V3} series GPU-supported VMs to run the CARLA simulator and \TheName's Simulation Manager. Four Standard D64s v5 \cite{Dsv5} are used to run the Vehicle Clients. The clients are evenly distributed across all four VMs.

\subsubsection{Multi-Node with Perception Enabled}

We utilize 3 Standard NV12s v3 \cite{NVS_V3} series GPU-supported VMs to run vehicle clients. We use the Standard ND40rs v2~\cite{NdM} series VMs to run the CARLA simulator, \TheName's Simulation Manager and vehicle clients. This VM features eight Tesla V100 GPUs, one of which runs CARLA, and the rest run vehicle clients with perception enabled. This is the configuration that supports the largest simulation scale.  

\subsection{Single-Node Experiments}

\textbf{CARLA Server Scalability. } To analyze the scalability of state-of-the-art methods, we measure the scalability of the CARLA server running on a single node. We vary the number of vehicles simulated as they autonomously drive around Town06. The AVs are spawned in random reproducible locations in the Town06 map and use direct data from the simulation to move around. 

\begin{figure}[htbp]
    \centering
    \includegraphics[width=0.5\textwidth]{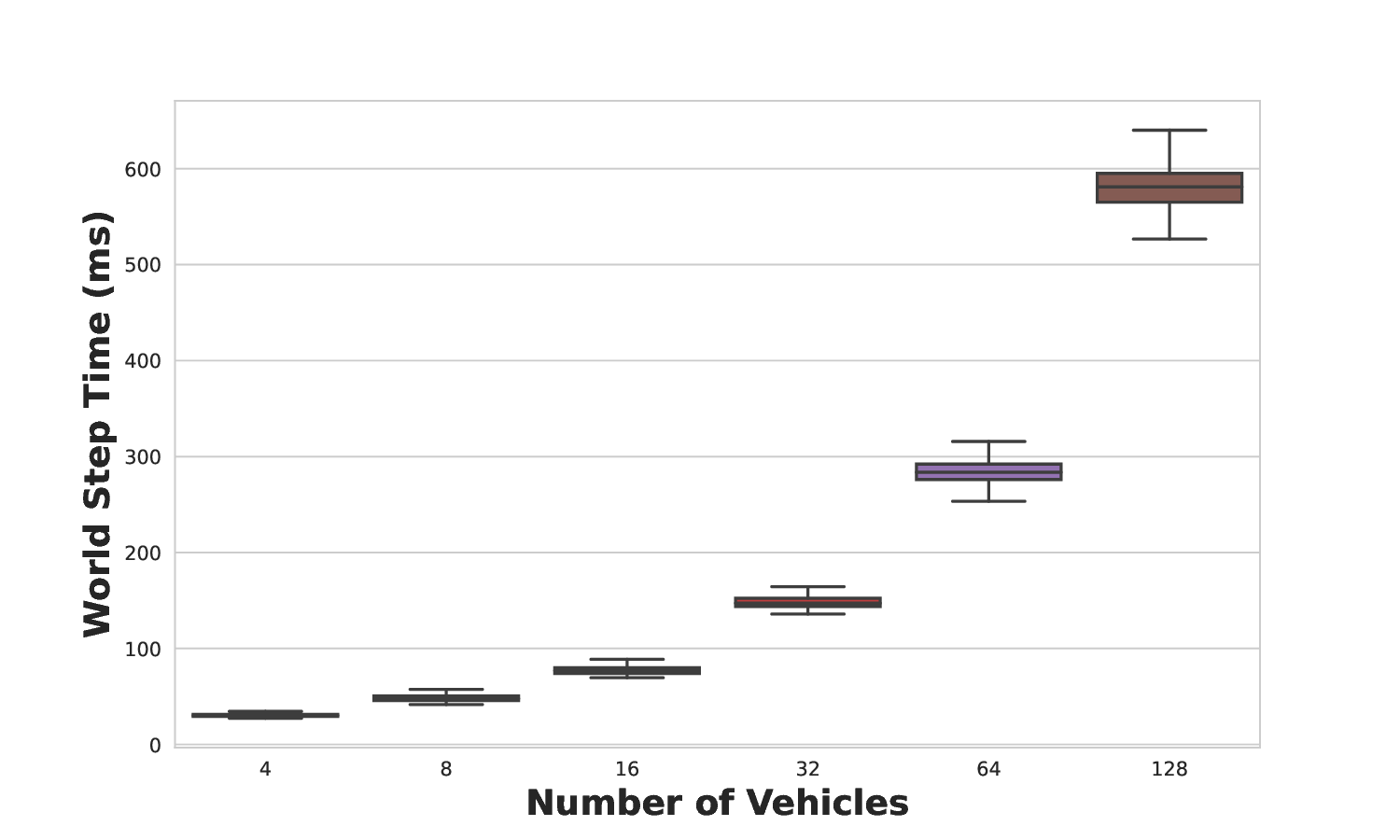}
    \vspace{-8mm}
    \caption{CARLA World Step Time}
    \Description[]{}
    \label{simtime_single_noperception}
\end{figure}

We first measure the world step time, which is a simulation component that remains unchanged between the baseline system and \TheName. 
Figure \ref{simtime_single_noperception} shows that the CARLA server's performance drops with an increasing number of cars.  As a component of total simulation time, we find that the world step time is a significant component, at least half of the total simulation tick time, even in a sequential scenario.


\textbf{Scalability Comparison. } We now compare the scalability of \TheName to the state-of-the-art OpenCDA~\cite{xu2021opencda}  simulation framework as a function of simulated AVs, using a single physical compute node. 
The experiment spawns vehicles in random reproducible locations in the Town06 map.

\begin{figure}[htbp]
    \centering
    \includegraphics[width=\columnwidth]{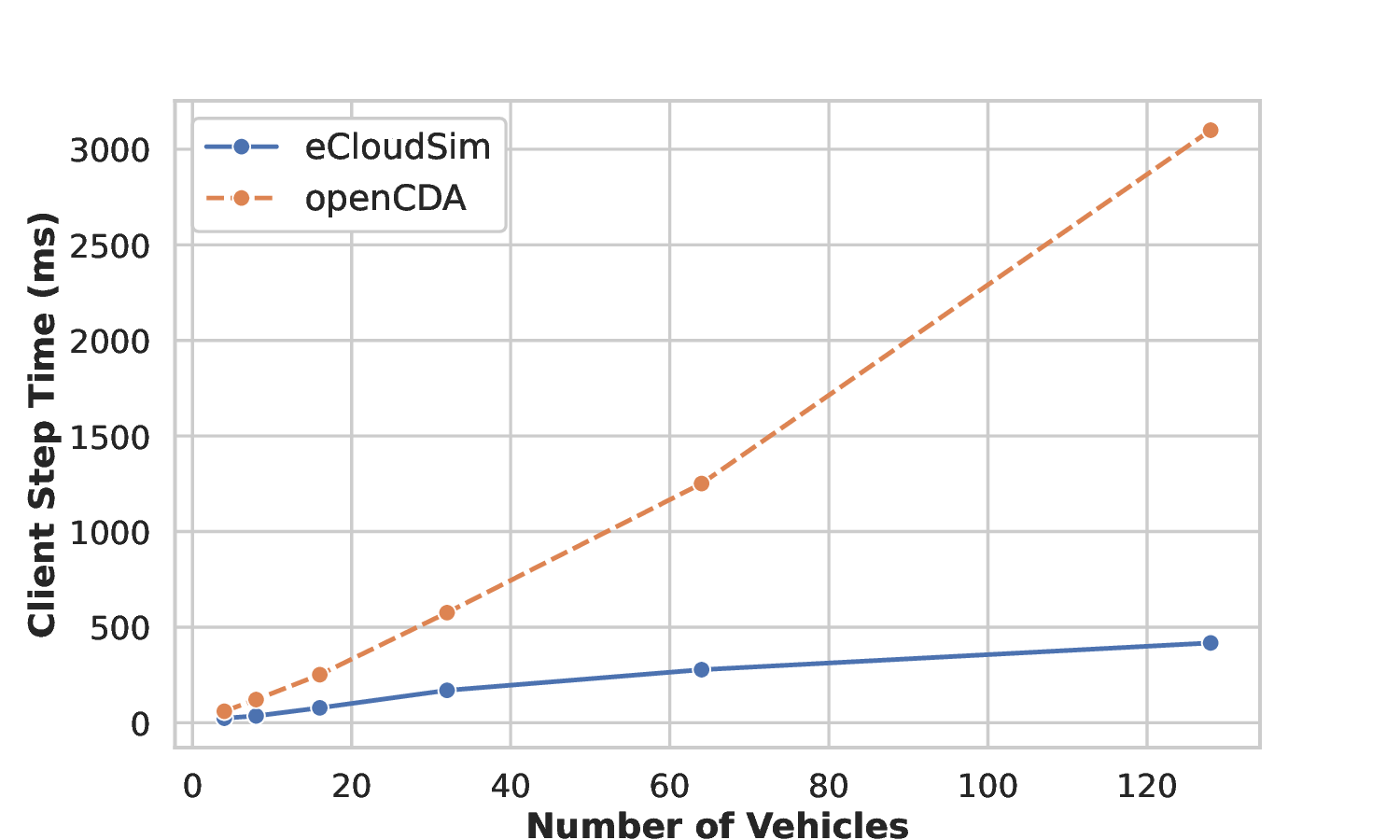}
    \vspace{-8mm}
    \caption{Total Client Step Time on a Single Node without Perception Enabled}
    \label{client_step_time_single_opencde_ecloudsim_no_perception}
\end{figure}

\begin{figure}[htbp]
    \centering
    \includegraphics[width=\columnwidth]{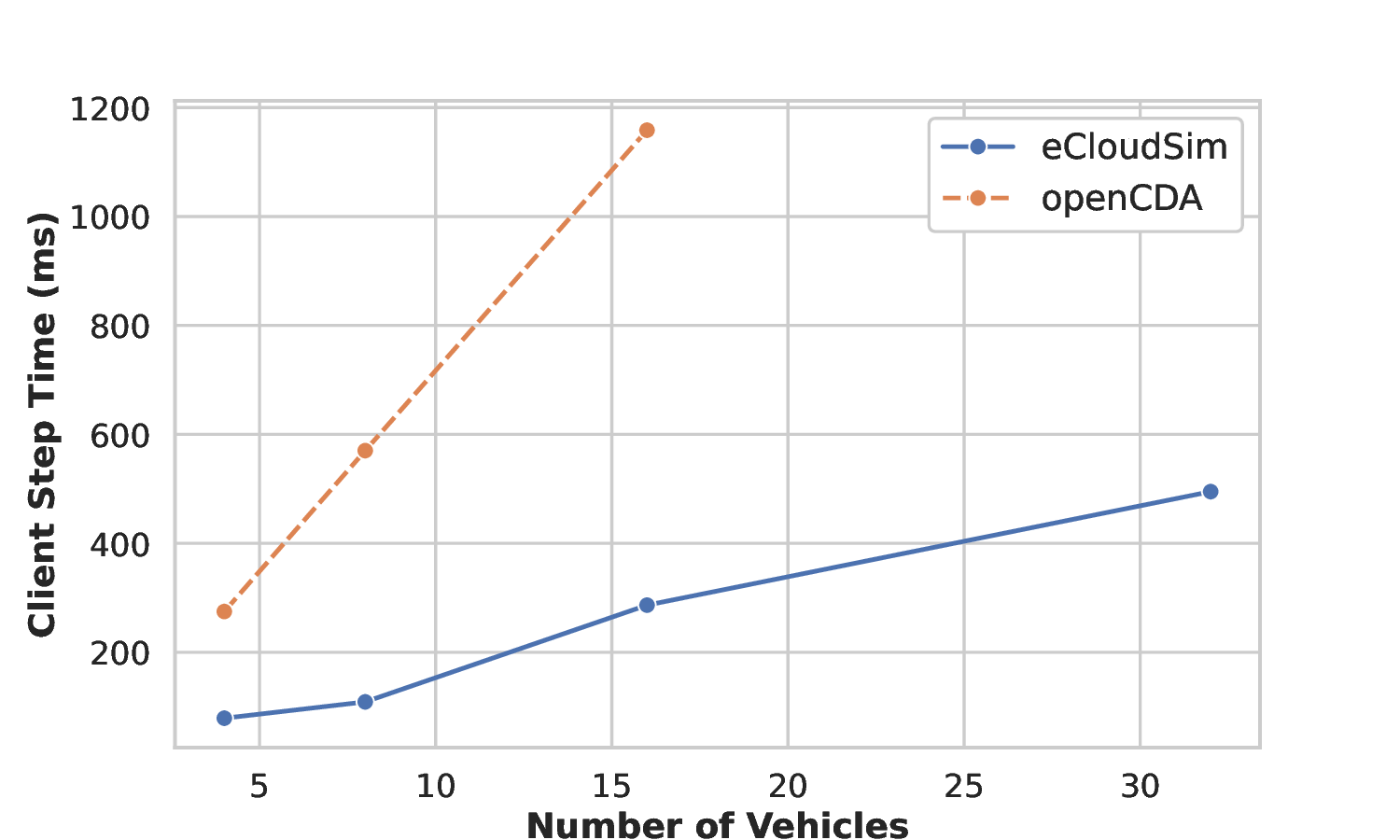}
    \vspace{-8mm}
    \caption{Total Client Step Time on a Single Node with Perception Enabled}
    \Description[]{}
    \label{client_step_time_single_opencda_ecloudsim_perception}
\end{figure}

Figure \ref{client_step_time_single_opencde_ecloudsim_no_perception} shows results without perception enabled.
\TheName is significantly more scalable than openCDA because it parallelizes the clients on the available cores, while the baseline runs all clients serially.
Beyond 128 cars, the simulation slows down significantly because the number of vehicles clients exceeds the number of hardware threads available for embarrassingly parallel computation.

Perception exacerbates the scalability challenge. Figure \ref{client_step_time_single_opencda_ecloudsim_perception} shows that the simulation time becomes untenable with only 16 cars because of OpenCDA's limitation of single GPU usage. Even with a powerful GPU, each vehicle requires about 1.5GB of GPU memory. 
Hence, the ability to scale out to multiple GPUs and compute nodes is essential for scalable simulations, and is lacking from state-of-the-art simulators. On a single node with 8 available GPUs, the simulation should theoretically be able to scale to 8x the vehicles. However, due to GPU memory constraints of the CARLA simulator and lack of multi-GPU support in CARLA 0.9.12, \TheName can double the number of simulated perception-enabled vehicles to 32, achieving a client step time that is only $\sim40\%$ of openCDA's achieved step time when simulating only half the AVs. 

\begin{figure}[htbp]
    \centering
    \includegraphics[width=\columnwidth]{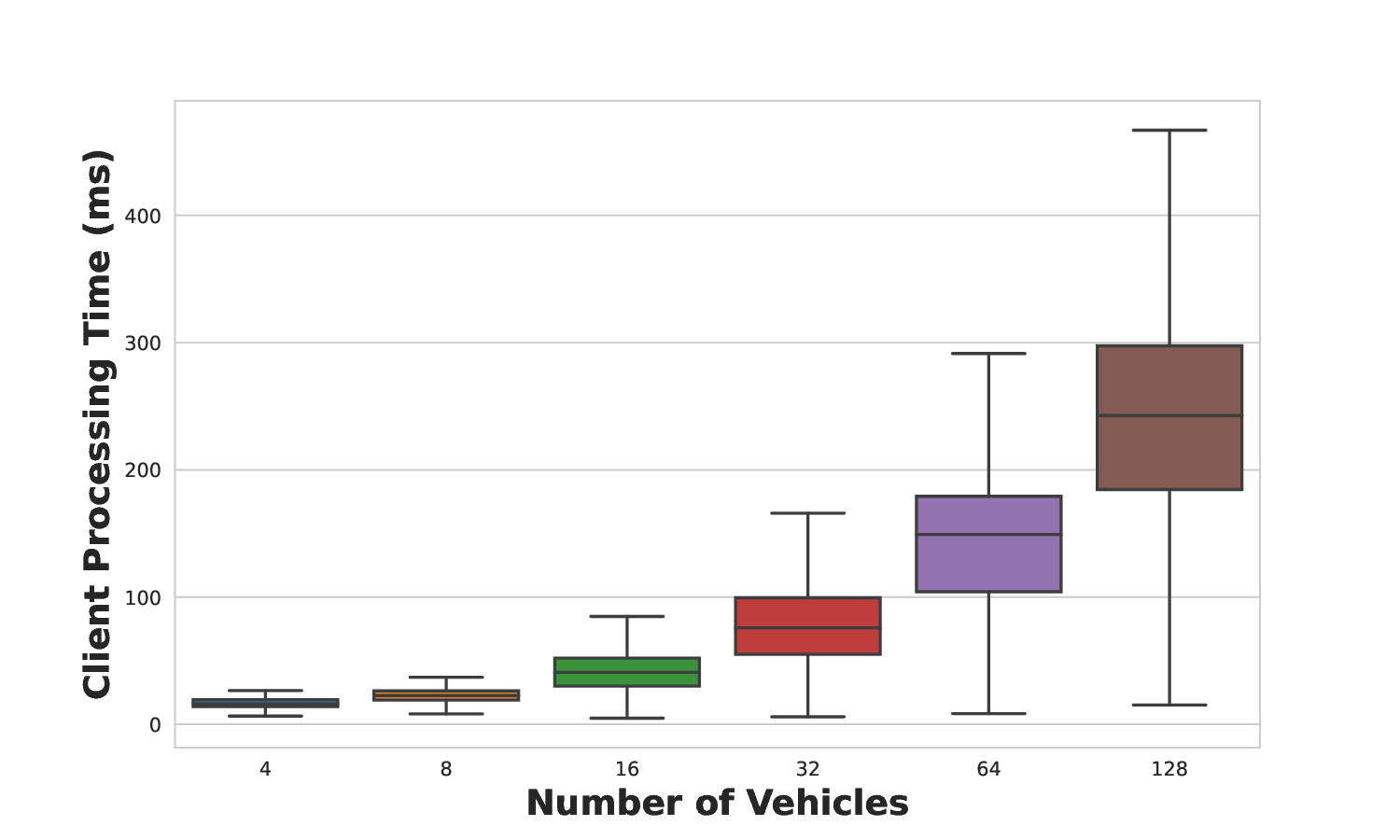}
    \vspace{-8mm}
    \caption{Client Processing Time - Individual Client Internal Control Logic without Perception Enabled}
    \label{client_procesing_time}
\end{figure}

Since \TheName fully parallelizes client processing, client step time---unlike world step time (Figure \ref{client_step_time_single_opencde_ecloudsim_no_perception})---should ideally remain constant as the number of client grows.
However, client step time gradually increases due to two factors: (i) as the number of clients grows, individual per-client processing requirements grow, and (ii) our parallelization introduces communication and synchronization overheads between CARLA and the clients.
We further analyze both of these effects next.
Overall, client step time consists of processing time, communication time, and barrier time, and its overall value is ultimately dictated by the \textit{slowest} client, because a barrier synchronization is required before the following world step can start.

Figure \ref{client_procesing_time} shows that the client processing time increases with the number of cars. Cars occasionally need to re-plan their course, which takes extra time, and the occurrence of such events increases with the number of cars.

\begin{figure}[htbp]
    \centering
    \includegraphics[width=\columnwidth]{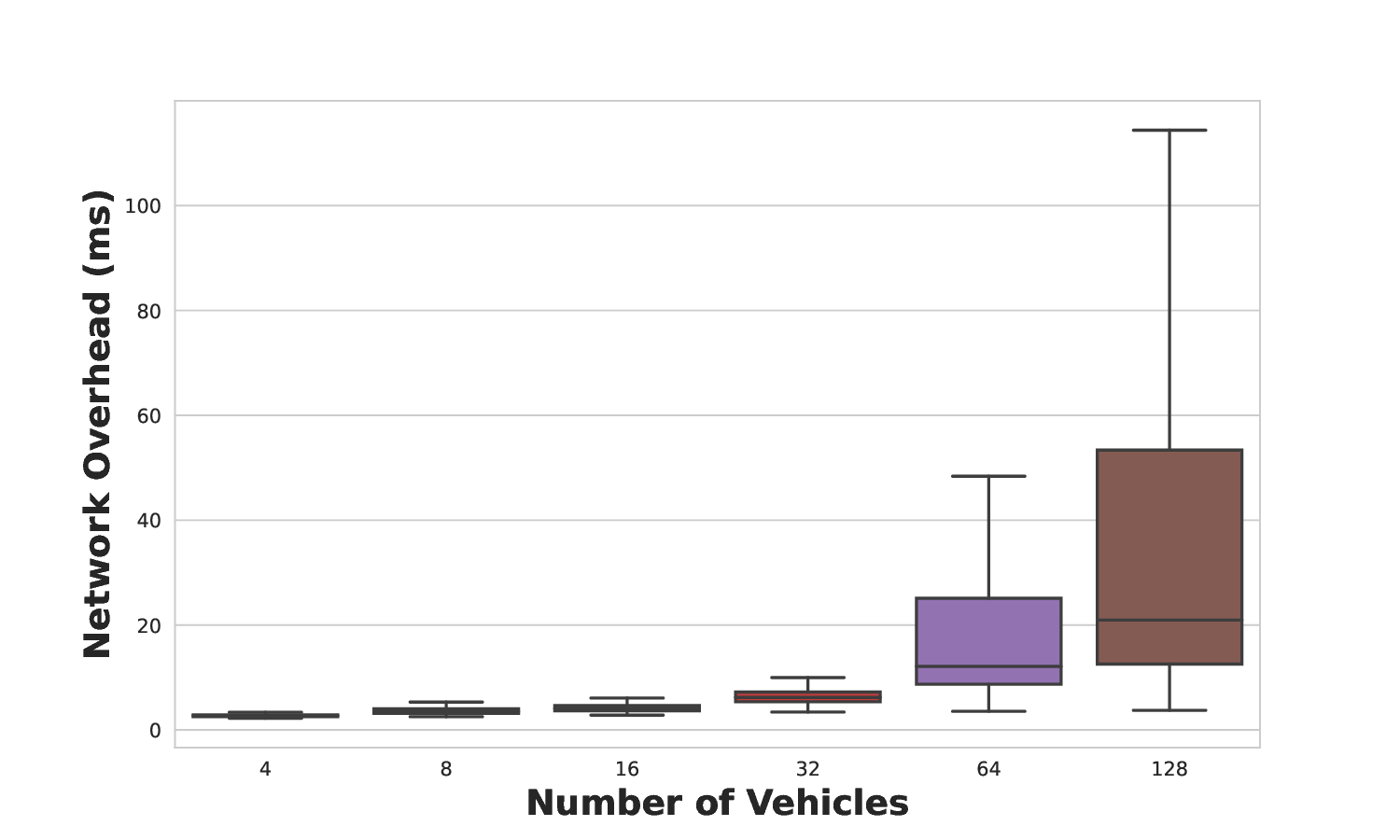}
    \vspace{-8mm}
    \caption{Network Overhead Time - Network Latency including Round Trip Time and Messaging Overhead without Perception Enabled}
    \Description[]{}
    \label{ecloudsim_network_time}
\end{figure}

\begin{figure}
    \centering
    \includegraphics[width=\columnwidth]{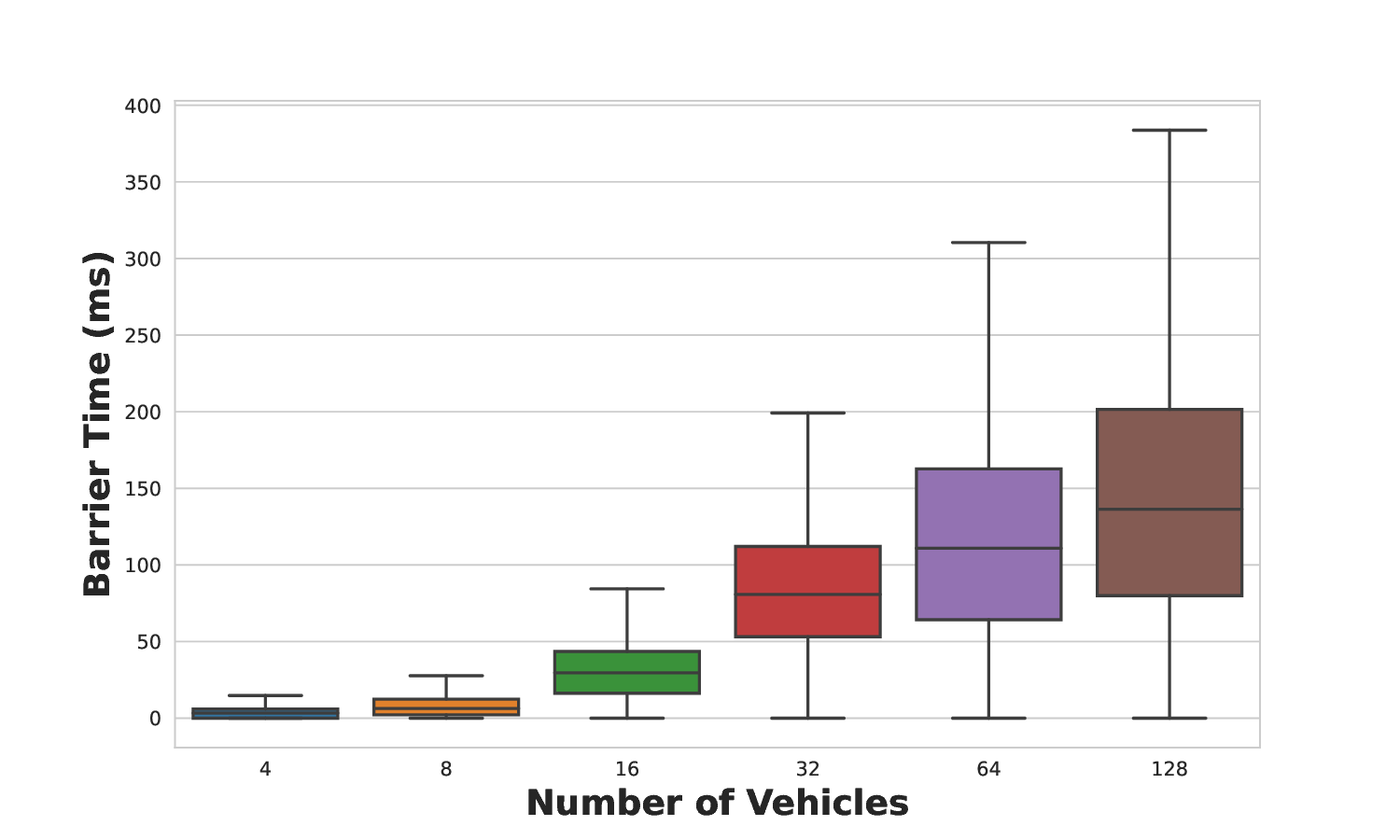}
    \vspace{-8mm}
    \caption{Time at Barrier - Idle Time at Barrier Waiting for All Clients to Complete Processing Logic without Perception Enabled}
   \Description[]{}
    \label{ecloudsim_barrier_time}
\end{figure}

Figures \ref{ecloudsim_network_time} and  \ref{ecloudsim_barrier_time} show the overhead \TheName introduces to parallelize and distribute the simulation. Compared to processing time, the network overhead's contribution to total client step time is small. However, barrier synchronization time increases considerably with the number of vehicles because it takes longer for at least one client to complete its processing logic computations. Removing the frequency of barrier synchronization would improve client step time, at the cost of simulation accuracy loss. 
The communication and synchronization overhead added by \TheName is consistent regardless of perception being enabled and does not increase until the system runs out of resources. 

\TheName is more scalable than previous simulators, even on a single node, because it bypasses the serialization overheads of the Python GIL and can therefore be parallelized across all available CPU resources. In sequential simulators, a single thread is a limiting factor for the simulation time.



\subsection{Multiple Node Experiments}

In these experiments, we evaluate the performance and scalability of \TheName alone, as no other simulation framework can scale to multiple nodes. We explore the limitations of our system as they relate to scalability and performance.

\subsubsection{Client Step Time}

\begin{figure}[htbp]
    \centering
    \includegraphics[width=\columnwidth]{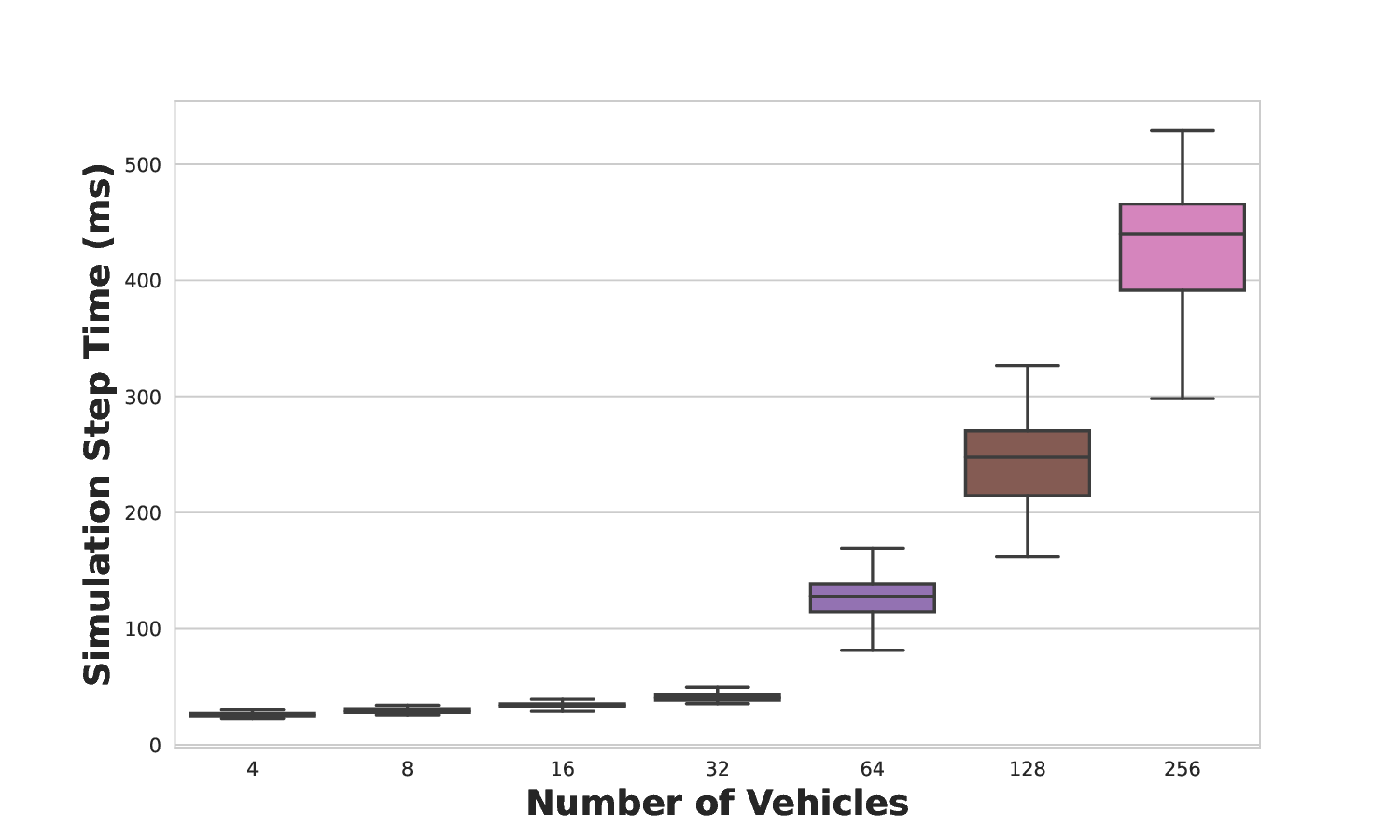}
    \vspace{-6mm}
    \caption{Multi Node Client Step Time without Perception Enabled}
   \Description[]{}
    \label{eval_multi_node_steptime_noperception}
\end{figure}

\begin{figure}
    \centering
    \includegraphics[width=\columnwidth]{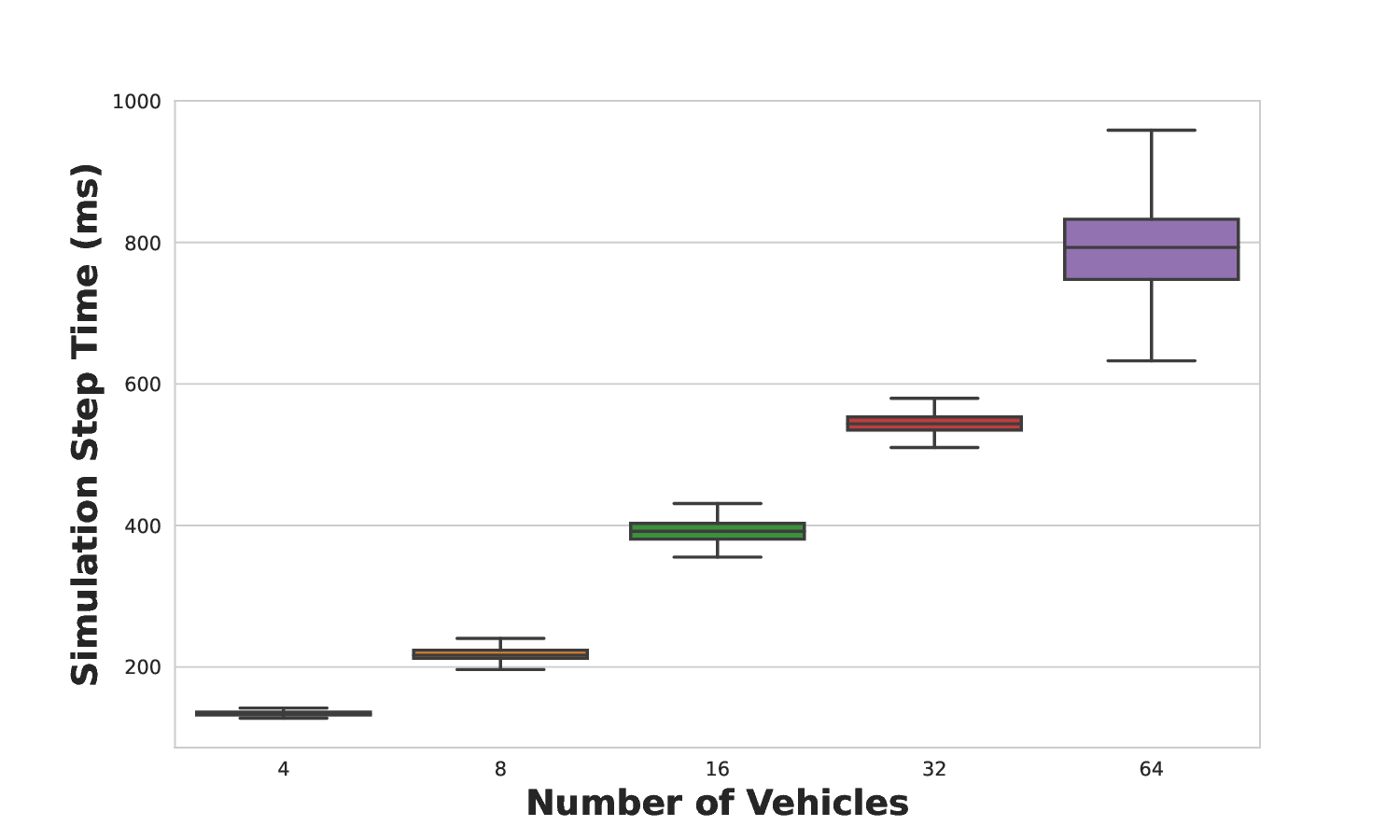}
    \vspace{-6mm}
    \caption{Multi Node Client Step Time with Perception Enabled}
   \Description[]{}\textbf{}
    \label{eval_multi_node_steptime_perception}
\end{figure}

Figures \ref{eval_multi_node_steptime_noperception} and \ref{eval_multi_node_steptime_perception} show the average client step time without and with perception, respectively. 
Client step time increases logarithmically with the number of clients,  due to a combination of resource contention on the machines themselves and the fact that the client processing time increases as the number of vehicles increases (as explained earlier). The more vehicles in a given scenario, the higher the probability of a slow vehicle due to a high- complexity decision that must be made in the increasingly busier environment. 
We did not identify a scalability bound in the infrastructure we developed; the network overhead incurred in the distributed scenario including RTT is only 45ms at the 99th percentile with 256 vehicle clients.
However, CARLA will currently not support more than ~300 vehicle actors due to a memory leak.

Comparing Figures \ref{eval_multi_node_steptime_noperception} and \ref{client_step_time_single_opencde_ecloudsim_no_perception}, we can see that \TheName can simulate a single step with 256 individual vehicles (without perception) faster than the state of the art can run a simulation with 32 vehicles. \TheName can run 128 vehicles approximately $6\times$ faster than the state of the art.

With perception enabled (Figure \ref{eval_multi_node_steptime_perception}), we see a similar client step time scaling trend, with over 80\% of the increase due to client processing time increase. Network overhead incurred by \TheName at the 99th percentile is only 35ms with 64 vehicle clients. The rest is due to uneven client processing. The perception evaluation is limited to 64 vehicles due to the CARLA actor limitation.
Since sensors are also spawned in perception-enabled simulation---four extra actors are created per vehicle, RGB Cameras (x3) and LiDAR---the memory limit for actor count is reached significantly sooner. 

\subsection{Accuracy}

To verify accuracy of the simulation, we compare individual vehicle client data from a reproducible simulation between \TheName and OpenCDA. These simulations should have the same vehicle clients performing the same maneuvers with the same data. 

\begin{figure}
    \centering
     \begin{subfigure}[b]{\columnwidth}
         \includegraphics[width=\columnwidth]{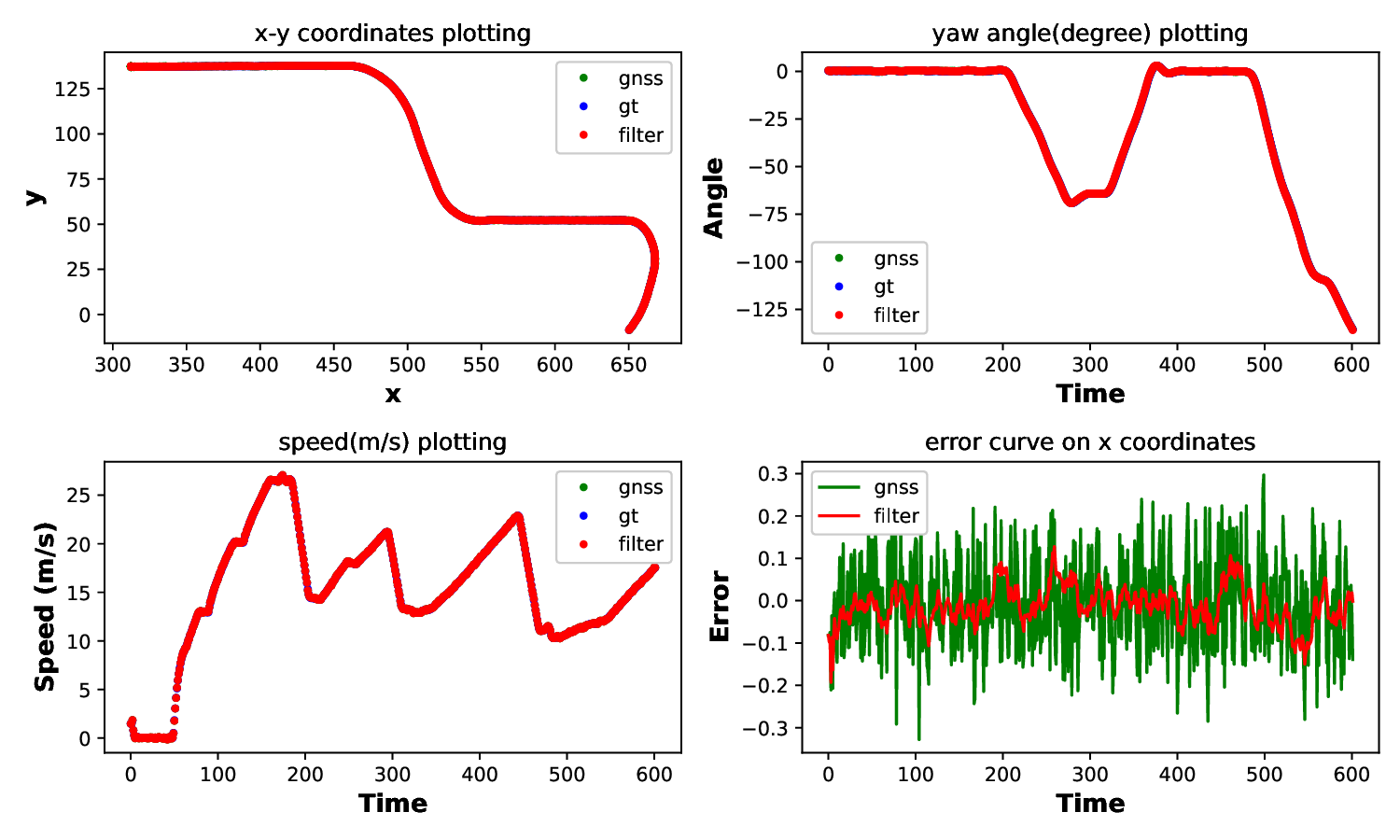}
         \vspace{-6mm}
         \caption{\TheName Localization Data}
         \Description[]{}
         \label{eCloudSim_localization}
     \end{subfigure}
     \begin{subfigure}[b]{\columnwidth}
         \includegraphics[width=\columnwidth]{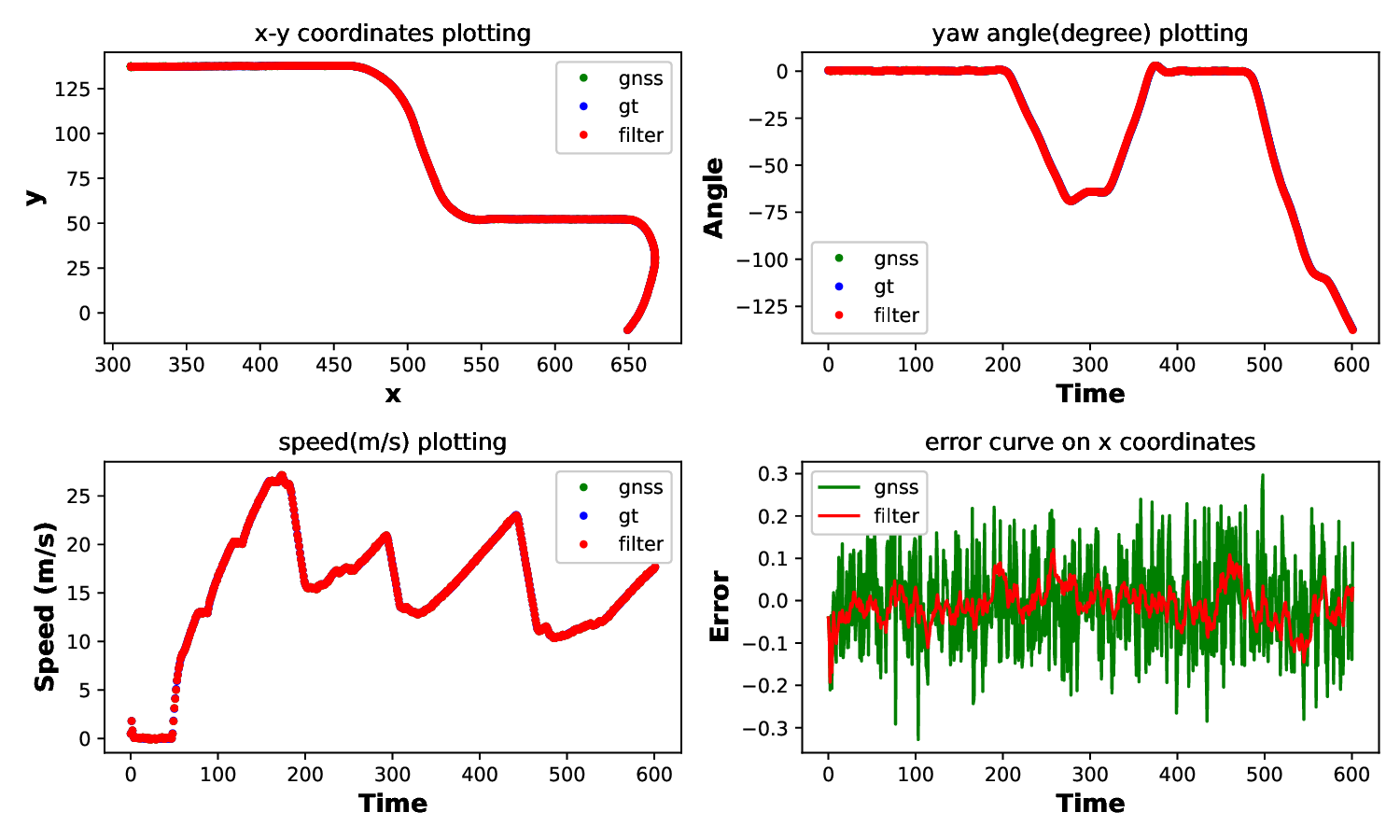}
         \vspace{-4mm}
         \caption{OpenCDA Localization Data}
         \label{opencda_localization}
     \end{subfigure}
    \caption{Comparison between OpenCDA and \TheName Ground Truth Data - Localization}
    \Description[]{}
    \label{localization_comparison}
\end{figure}

\cref{localization_comparison} shows localization data for both \TheName and OpenCDA for the same scenario.
We show plots for four metrics: the x-y coordinate planning (i.e., location of the vehicle), yaw angle, speed, and an error measurement comparing GNSS measurement to ground truth data through a filter. 
We observe that the figures are practically identical, indicating that \TheName preserves the simulation accuracy of OpenCDA.

\subsection{Usability}
\label{sec:eval:usability}

Besides scalability, usability is a key qualitative goal for \TheName. To demonstrate usability, we developed an exemplar algorithm to run in the edge and assist in collaborative maneuvering on a simple four lane highway. 

\subsubsection{Edge Integration}

\label{sec:algo}
To simulate different edge-assisted CAV scenarios, \TheName allows deployment of different pluggable algorithms on the edge node for improvement in traffic flow and vehicle safety. The modular edge node holds the algorithm functional model for each algorithm to run on the edge. The below sections describe a currently deployed functional, edge-based algorithm that has been deployed and tested in \TheName.

Each vehicle in the algorithm is modeled using four states: (a) linear position coordinate on the road (measured in meters), (b) lane position (in terms of lane number), (c) velocity (in meters per second (m/s)), (d) target velocity desired by the vehicle (in m/s). 
We employ two different algorithmic approaches for traffic management as concrete case studies:  
\begin{enumerate}
\item
\textbf{Edge-based}: A system that assigns lanes to vehicles based on target velocities, clustering vehicles based on their intentions and enabling faster traffic flow; and 
\item 
\textbf{Greedy}: A non-collaborative approach wherein each vehicle changes lanes and moves to reach its target velocity with no outside input or communication.
\end{enumerate}
In both algorithms, each vehicle can accelerate or decelerate by 1m/s at each time step, or can change lanes. These simplified dynamics are used to plan and reevaluate the path of each vehicle at every simulation time step.
 
The centralized approach is implemented as a graph search, where vehicles search through the discretized space of control inputs (velocity and lane change commands) to find the optimal combination of inputs in each group of vehicles that minimizes deviation from target velocity. 
We use A* search \cite{Nilsson_astar_ai} with node states represented as the vector $[v_1,l_1,...,v_N,l_N]^T$, where $v_i$ is the $i^{th}$ vehicle's velocity and $l_i$ is its lane. Velocity changes and lane change commands are used to move through the search space, constrained by the need to avoid collisions between vehicles while staying in their lane. Each node also carries the attendant vehicle position coordinates (lane and linear position), allowing the graph search to check for and avoid collisions (constraint on the search). Due to the need to have two states per vehicle and due to the $A^*$ search's exponential complexity, this approach can deliver asymptotically optimal results but is difficult to scale to more than a few vehicles at a time.

To limit the effects of $A^*$'s complexity, we use constrained k-means clustering \cite{constrained_kmeans} to group vehicles based on the distance between them, splitting the vehicles into groups of a fixed maximum size (in this test case, up to 3 vehicles per cluster). This approach runs clustering each time step to ensure that changes in each vehicle's target velocity is accounted for. The clustering is done using target velocities sent by each vehicle to an edge node, and the edge node then sends back suggested lanes to each vehicle. Each vehicle moves to reach its suggested lane as and when possible, with no explicit centralized coordination. Each cluster plans its path based on the current positions of the other clusters, as opposed to the centralized approach where each vehicle accounts for the future (planned) positions of other vehicles.

The greedy approach is implemented with each vehicle attempting to reach its target velocity in its current lane and changing lanes only if that is no longer possible (e.g., due to a slower vehicle blocking the path). 
There is neither centralized coordination nor edge-based suggestions in this approach; the only assumption is that each vehicle is able to guesstimate the velocity of the vehicles in its line of sight, which can lead to collisions or slower traffic flow.

\subsubsection{Edge-Vehicle Client Control Learnings}

\par We conducted an experiment in which we forcefully overrode the waypoint buffer in the vehicle client used for local path planning with waypoints calculated from the A* algorithm used above. We found that the cars failed to meet control requirements at the network latency used in the experiment (51ms in our simple experiment), which would also occur in a real-world system. To address this issue, we extended the search time horizon for the edge to account for the increase in communication latency.  
This simple experiment uncovers a useful finding about integrating edge-based control into vehicles: to improve vehicle dynamics, the search time horizon for the edge control plane must take into account the staleness of information due to communication latency. 

\subsubsection{Edge Algorithm Statistics Gathering}

We now use an example scenario to compare the two different types of AV control mentioned in Section \ref{sec:algo} (greedy and edge-based), and demonstrate statistics that can be collected from the simulation.

\textbf{Traffic flow} can be measured using the deviation from target velocity.
    
\textbf{Safety} can be measured in safety rule violations, minimum headway distance violations between vehicles in same lane, vehicle traversal outside of lane, and car blockage (two cars stopped, facing each other with no way around). Safety is generally the most important metric for an autonomous driving system. We measure safety using data collected by CARLA for these specific metrics that can be accessed through the simulation manager's evaluation interface.


\textbf{Systems Metrics} such as latency, computational time, and latency requirements violations for a control loop of 5Hz (200ms period) end-to-end should be considered. These can be accessed and extended via the edge evaluation manager.

\begin{figure}
    \centering
     \begin{subfigure}[b]{\columnwidth}
         \includegraphics[width=\columnwidth]{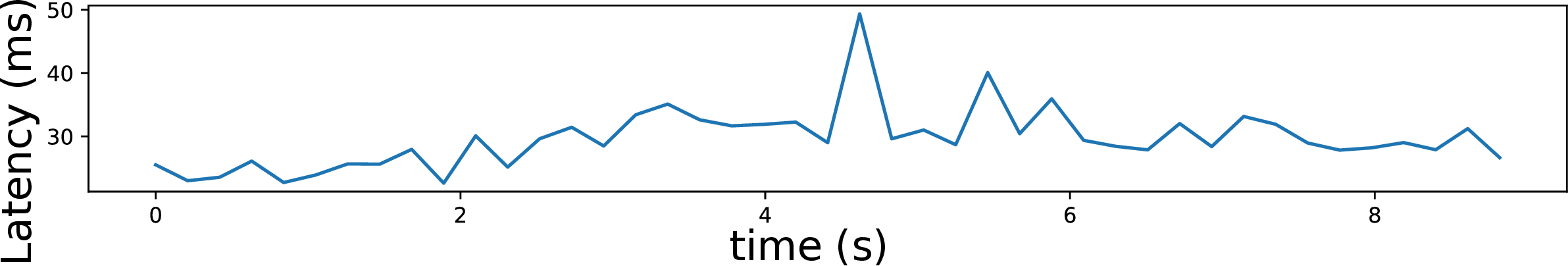}
         \caption{A* Algorithm Execution Time - 4 Cars}
        \Description[]{}
         \label{edge_mean_algorithm_latency_4}
     \end{subfigure}
     \begin{subfigure}[b]{\columnwidth}
        \centering
        \includegraphics[width=\columnwidth]{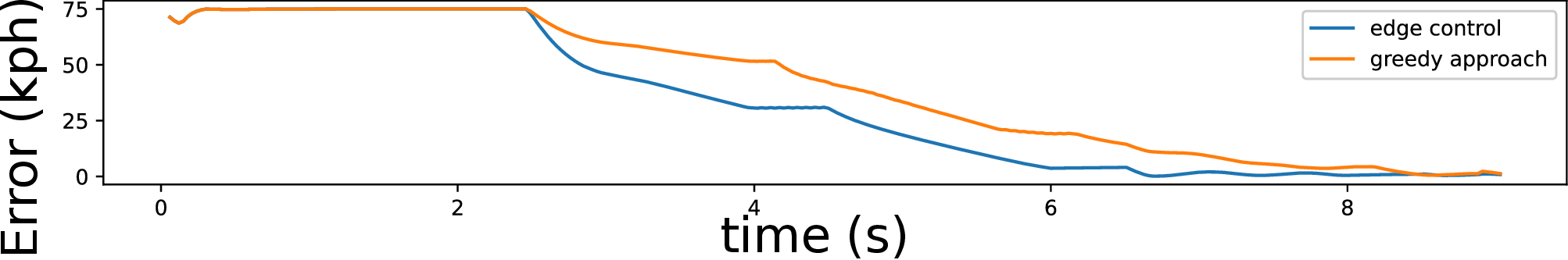}
        \caption{Deviation From Target Velocity vs Simulation Time - Edge-Assisted Control Versus per-AV Greedy Control}
        \label{ego_edge_velocity_error}
     \end{subfigure}
        \caption{Traffic and Algorithm Time Metrics Gathered from \TheName}
        \Description[]{}
        \label{eval_edge_mean_algorithm_latency_graphs}
\end{figure}

\par The \texttt{edge\_debug\_helper} class provides an extensible interface for gathering statistics on the edge. Similar to other statistics gathering classes on clients and the simulator, it allows us to glean important information about the performance of the edge algorithm in real time.

\par Figure \ref{edge_mean_algorithm_latency_4} shows the time it takes to run the edge algorithm depending on the number of cars it owns, and compares the A* algorithm calculation time for a subset of vehicles owned by a single edge node over time. We show the distribution of calculation times required at the edge (using a single-threaded A* implementation), and the places where the edge takes the longest time to complete the calculations.

\par Figure \ref{ego_edge_velocity_error} shows the deviation from target velocity over time. This simple experiment demonstrates the promise of edge-assisted CAV: the edge-based control plane was able to help the vehicles reach their target speed of 75kph more quickly due to centralized coordination of the vehicle clients. For a more challenging scenario in which four vehicles were spawned in close proximity to each other, the edge-based control plane achieved a more than 15\% improvement in mean vehicle velocity (42kph vs 35kph). We omit the critically important metric of collisions, as there were none observed in any of these exemplar scenarios. 

Not that, in this experiment, we did not attempt to highly optimize the performance of the A* algorithm or evaluate the quality of different possible edge-based algorithms.
Instead, the experiment's purpose is to demonstrate that \TheName can be used to set up and evaluate different CAV scenarios, including futuristic systems with edge support.
\section{Conclusion}

We have presented a scalable and distributed platform for Connected Autonomous Vehicles (CAVs) evaluation, called \TheName, which will be open-sourced to facilitate CAV research.  
\TheName can emulate large-scale CAV scenarios with several dozens to hundreds of AVs (with and without perception enabled), at least $5\times$ faster than prior art through scalable deployment on a cluster of machines or the cloud. It also supports simulation for the edge as a broker between vehicles, opening new avenues for large-scale edge-assisted CAV research. Our future directions to improve the utility and scope of \TheName include:

\begin{enumerate}
    \item Space sharing identical perception models to allow for larger simulation scenarios with fewer resources.
    \item Multi-edge support for larger geographical area to expand the scope of CAV research for metropolitan area CAV instances.
\end{enumerate}


\bibliographystyle{ACM-Reference-Format}
\bibliography{references.bib}
\end{document}